\documentclass[11pt]{article}

\usepackage[final]{acl}

\usepackage{times}
\usepackage{enumitem}
\usepackage{latexsym}
\usepackage[T1]{fontenc}
\usepackage[utf8]{inputenc}
\usepackage{microtype}
\usepackage{inconsolata}
\usepackage{graphicx}
\usepackage{booktabs}
\usepackage{multirow}
\usepackage{xcolor}
\usepackage{amsmath,amssymb}
\usepackage{comment}
\usepackage{listings}
\usepackage{tcolorbox} 
\usepackage{authblk}
\usepackage{graphicx}
\usepackage{forest}
\useforestlibrary{edges}
\usepackage{xcolor}

\usepackage{rotating}
\usepackage{natbib}
\usepackage{url}
\usepackage{fontawesome5}
\tcbuselibrary{breakable, skins}
\definecolor{promptbackground}{RGB}{245, 245, 245}
\definecolor{promptborder}{RGB}{220, 220, 220}
\definecolor{l1blue}{HTML}{4A90C4}
\definecolor{l1bluefill}{HTML}{DCEEFB}
\definecolor{l1green}{HTML}{4AAF4A}
\definecolor{l1greenfill}{HTML}{DEF2DE}
\definecolor{l1yellow}{HTML}{D4A017}
\definecolor{l1yellowfill}{HTML}{FFF6D5}
\definecolor{l1purple}{HTML}{8E6BBF}
\definecolor{l1purplefill}{HTML}{EDE0FA}

\definecolor{rootfill}{RGB}{240, 240, 240}
\definecolor{l1blue}{RGB}{0, 102, 204}
\definecolor{l1bluefill}{RGB}{230, 242, 255}
\definecolor{l1green}{RGB}{0, 153, 76}
\definecolor{l1greenfill}{RGB}{230, 255, 238}
\definecolor{l1yellow}{RGB}{204, 153, 0}
\definecolor{l1yellowfill}{RGB}{255, 250, 230}
\definecolor{l1purple}{RGB}{102, 0, 204}
\definecolor{l1purplefill}{RGB}{242, 230, 255}

\definecolor{rootfill}{HTML}{EAEAEA}
\definecolor{leaf}{HTML}{333333}

\newtcolorbox{promptbox}{
    colback=promptbackground,
    colframe=promptborder,
    sharp corners,
    boxrule=0.5pt,
    fontupper=\small\ttfamily, 
    left=10pt,
    right=10pt,
    top=10pt,
    bottom=10pt
}
\title{Anatomy of Agentic Memory: Taxonomy and Empirical Analysis of Evaluation and System Limitations}

\author{Dongming Jiang$^{\alpha}$, Yi Li$^{\alpha}$, Songtao Wei$^{\alpha}$, Jinxin Yang$^{\alpha}$, Ayushi Kishore$^{\beta}$, Alysa Zhao$^{\gamma}$,
Dingyi Kang$^{\alpha}$, \\ Xu Hu$^{\alpha}$, Feng Chen$^{\alpha}$, Qiannan Li$^{\beta}$ and Bingzhe Li$^{\alpha,\ast}$\\
$^{\alpha}$University of Texas at Dallas \quad $^{\beta}$University of California, Davis \quad $^{\gamma}$Texas A\&M University\\

{\footnotesize \{dongming.jiang, yi.li3, songtao.wei, jinxin.yang, dingyi.kang, xu.hu, feng.chen, bingzhe.li\}@utdallas.edu}\\
{\footnotesize \{aykishore, qnli\}@ucdavis.edu; alysazhao111@tamu.edu}\\
{\footnotesize $^\ast$Corresponding author \quad \href{https://github.com/FredJiang0324/Anatomy-of-Agentic-Memory}{\faGithub~GitHub Repo}}}
\begin{document}
\maketitle

\begin{abstract}
Agentic memory systems enable large language model (LLM) agents to maintain state across long interactions, supporting long-horizon reasoning and personalization beyond fixed context windows. Despite rapid architectural development, the empirical foundations of these systems remain fragile: existing benchmarks are often underscaled, evaluation metrics are misaligned with semantic utility, performance varies significantly across backbone models, and system-level costs are frequently overlooked. This survey presents a structured analysis of agentic memory from both architectural and system perspectives. We first introduce a concise taxonomy of Memory-Augmented Generation (MAG) systems based on four memory structures. Then, we analyze key pain points limiting current systems, including benchmark saturation effects, metric validity and judge sensitivity, backbone-dependent accuracy, and the latency and throughput overhead introduced by memory maintenance. By connecting the memory structure to empirical limitations, this survey clarifies why current agentic memory systems often underperform their theoretical promise and outlines directions for more reliable evaluation and scalable system design. To facilitate future research, we maintain an open-source repository of surveyed papers, benchmarks, and resources.\footnote{\url{https://github.com/FredJiang0324/Anatomy-of-Agentic-Memory}. Given the rapid evolution of research in agentic memory, we may have inadvertently overlooked some relevant studies. We welcome you to suggest missing papers by emailing us or opening an issue on our GitHub repository.}
\end{abstract}

% ==================================================
\section{Introduction}
Large language model (LLM) agents are increasingly expected to operate over long time horizons, maintaining user preferences, accumulating task-relevant knowledge, etc.~\cite{Brown2020, achiam2023gpt, Wei2022}. However, fixed context windows fundamentally limit their ability to retain and manipulate persistent state~\cite{Brown2020, Beltagy2020, liu2024lost, Press2021}. To address this constraint, Memory-Augmented Generation (MAG) extends memory beyond the prompt, enabling agents to store, retrieve, and update information across interactions~\cite{xu2025mem, nan2025nemori, chhikara2025mem0, jiang2026magma, liu2026simplemem}. While this paradigm has rapidly evolved from lightweight semantic stores to entity-centric, reflective, and hierarchical designs, empirical understanding remains limited: reported gains are inconsistent across benchmarks, highly backbone-dependent, and lack principled guidance on evaluation and system-level cost.

These challenges stem in part from underspecified design trade-offs and inadequate evaluation. Benchmarks are often underscaled relative to modern context windows, metrics emphasize surface overlap over semantic utility, and system-level costs such as latency and throughput degradation are rarely measured. As a result, complex memory systems are frequently tested in settings where simpler full-context or retrieval baselines suffice, obscuring their true benefits and limitations.

\begin{table*}[!t]
\centering
%\small
\caption{Comparison with related surveys on memory for LLM-based agents. \checkmark\ indicates the topic is systematically discussed; (\checkmark) indicates partial or brief coverage; \texttimes\ indicates the topic is not addressed.}
\label{tab:survey_comparison}
\setlength{\tabcolsep}{6pt}
\renewcommand{\arraystretch}{1.4}
\resizebox{\textwidth}{!}{%
\begin{tabular}{l | l | c c c c c}
\toprule
\textbf{Survey} & \textbf{Taxonomy Focus} & \shortstack{\textbf{Memory Mgmt.}\\\textbf{\& Policy}} & \shortstack{\textbf{Benchmark}\\\textbf{Saturation}} & \shortstack{\textbf{Metric}\\\textbf{Validity}} & \shortstack{\textbf{Backbone}\\\textbf{Sensitivity}} & \shortstack{\textbf{System Cost}\\\textbf{\& Latency}} \\
\midrule
\textbf{The AI Hippocampus}~\cite{jia2026ai} & Brain-inspired: implicit, explicit, agentic & (\checkmark) & \texttimes & \texttimes & \texttimes & \texttimes \\
\midrule
\textbf{Memory in the Age of AI Agents}~\cite{hu2025memory} & Forms--functions--dynamics & (\checkmark) & \texttimes & \texttimes & \texttimes & \texttimes \\
\midrule
\textbf{Toward Efficient Agents}~\cite{yang2026toward} & Efficiency-focused: memory, tool learning, planning & \checkmark & \texttimes & \texttimes & \texttimes & \checkmark \\
\midrule
\textbf{Rethinking Memory Mechanisms}~\cite{huang2026rethinking} & Substrate--cognition--subject & \checkmark & (\checkmark) & \texttimes & \texttimes & \texttimes \\
\midrule
\textbf{From Storage to Experience}~\cite{luo2026storage} & Evolutionary: storage--reflection--experience & (\checkmark) & \texttimes & \texttimes & \texttimes & \texttimes \\
\midrule
\textbf{Graph-based Agent Memory}~\cite{yang2026graph} & Graph-oriented lifecycle & \checkmark & (\checkmark) & \texttimes & \texttimes & \texttimes \\
\midrule
\textbf{Taxonomy and Empirical Analysis (Ours)} & Structural + Empirical analysis & \checkmark & \checkmark & \checkmark & \checkmark & \checkmark \\
\bottomrule
\end{tabular}%
}
\end{table*}

In this paper, we provide a structured analysis of agentic memory systems from both architectural and empirical perspectives. 1) We introduce a concise taxonomy of Memory-Augmented Generation organized around four memory structures: Lightweight Semantic, Entity-Centric and Personalized, Episodic and Reflective, and Structured and Hierarchical. Defined by how memory is organized and manipulated, this taxonomy establishes a principled basis for analyzing system behavior. 2) Building on this framework, we identify key bottlenecks limiting reliability and scalability, including benchmark saturation, metric misalignment (e.g., F1 vs. semantic correctness), prompt sensitivity in LLM-as-a-judge evaluation, backbone dependence, and system-level costs such as retrieval latency, update overhead, and throughput degradation.

%In this paper, we provide a structured analysis of agentic memory systems from both architectural and empirical perspectives. We first present a concise taxonomy of Memory-Augmented Generation architectures organized around four dominant memory structures: Lightweight Semantic Memory, Entity-Centric and Personalized Memory, Episodic and Reflective Memory, and Structured and Hierarchical Memory. The categories are defined by how memory is organized and manipulated. By isolating the structural primitives of agentic memory, the taxonomy provides a principled foundation for analyzing downstream behavior. 

%Second, building on this taxonomy, we identify the key bottlenecks limiting the reliability and scalability of agentic memory systems. We highlight benchmark shortcomings particularly context saturation, and show that common lexical metrics (e.g., F1) poorly capture semantic correctness, while LLM-as-a-judge evaluations remain sensitive to prompt design and calibration. We further examine backbone dependence as well as system-level costs such as retrieval latency, update overhead, and throughput degradation. 

By linking memory structures to empirical limitations, this survey clarifies why current agentic memory systems often fall short of their theoretical promise. Rather than advocating a single “best” design, we provide a diagnostic framework to explain when specific memory structures are effective, when they fail, and what trade-offs they entail. Our analysis offers guidance for designing more robust benchmarks, more reliable evaluation protocols, and more scalable agentic memory systems.

\textbf{Difference from other surveys:} While existing surveys~\cite{jia2026ai,hu2025memory,yang2026toward, yang2026graph, luo2026storage, huang2026rethinking} primarily operate at the \emph{theoretical} level by cataloguing architectures, defining conceptual taxonomies, and drawing cognitive science analogies, our survey bridges the gap \emph{from theory to practice}. Our taxonomy is structure-oriented, not only discussing various memory structure designs, but also highlighting the memory management frameworks and optimization strategies. In addition, we provide comprehensive evaluations across multiple benchmarks. Specifically, we conduct systematic analyses of agentic memory systems on benchmark saturation, metric validity, backbone sensitivity, and maintenance overhead, overlooked in prior surveys yet critical for understanding why current MAG systems often fall short of their theoretical promise. A detailed comparison is presented in Table~\ref{tab:survey_comparison}.%In addition, we take a further step towards a system-level examination through comprehensive evaluations across different benchmarks. In particular, we contribute systematic analyses of benchmark saturation, metric validity, backbone sensitivity, and maintenance overhead, which remain largely unexamined in prior surveys yet are critical for understanding why current MAG systems often fall short of their theoretical promise. A detailed comparison is listed in Table~\ref{tab:survey_comparison}.

\section{Background}\label{sec:background}
%To move beyond the limitations of purely in-context or retrieval-based approaches, recent work has increasingly adopted Memory-Augmented Generation (MAG), in which large language models interact with an external memory during generation \color{red} as shown in Figure xxx \color{black}. Early forms of MAG are exemplified by Retrieval-Augmented Generation (RAG), where relevant documents are retrieved from a static, read-only corpus and injected into the prompt. While effective for knowledge-intensive tasks, such designs treat memory as an external reference rather than a persistent state. Agentic memory systems generalize this paradigm by introducing writable, evolving memory that persists across interactions, enabling agents to store, update, and reuse information over long horizons. This shift from read-only retrieval to read–write memory fundamentally alters the role of memory, making it a core component of agent behavior rather than an auxiliary information source.

Agentic memory extends retrieval-based generation by introducing persistent, writable memory that evolves across interactions, enabling an LLM agent to store, update, and reuse information over time. Formally, at step $t$, the agent conditions on observations $o_t$ and an external memory state $\mathcal{M}_t$:
\begin{equation}
y_t \sim f_\theta\!\Big(\phi(o_t, s_t)\ \oplus\ \psi(\mathcal{M}_t; q_t)\Big),
\label{eq:agentic_memory_conditioning}
\end{equation}
where $y_t$ denotes the output, $s_t$ additional agent state, and $\psi(\mathcal{M}_t; q_t)$ retrieves memory given query $q_t$. The operator $\oplus$ represents integration (e.g., prompt concatenation or structured fusion). Crucially, memory affects behavior through the explicit retrieval term $\psi(\mathcal{M}_t; q_t)$ rather than updates to $\theta$.

Two coupled processes are operated in agentic memory: inference-time recall and memory update. At each step, the agent retrieves relevant information from an external memory store to condition its decision, and subsequently writes, updates, or consolidates memory to maintain a useful long-term state. Unlike parametric learning, this mechanism influences behavior through explicit read–write operations over an evolving memory state rather than by modifying model weights. A formalization of these operations including query generation, utility-aware retrieval, and memory actions such as store, summarize, link, and delete is provided in Appendix~\ref{sec:basic_behaviors}.

\begin{comment}
\paragraph{Lifecycle behaviors: formation, retrieval, consolidation, forgetting.}
Across implementations, external agentic memory exhibits recurring lifecycle behaviors:
\begin{itemize}
  \item \textbf{Formation (what to store):} selecting salient facts, preferences, commitments, tool outcomes, or environment observations from interaction.
  \item \textbf{Encoding (how to represent):} raw text/media, structured records (schemas/graphs), or embedding-based representations with metadata.
  \item \textbf{Indexing/Storage (where to store):} vector indices, hybrid dense+sparse indices, relational/graph stores, or episodic logs.
  \item \textbf{Retrieval/Injection (how to recall):} top-$k$ retrieval, multi-hop recall, planner-guided retrieval, and context packing/compression.
  \item \textbf{Update/Consolidation (how to evolve):} patching, merging, versioning, and consolidating episodes into stable summaries.
  \item \textbf{Forgetting/Auditing (how to control):} eviction or deletion (e.g., TTL, user-requested deletion), deduplication, provenance tracking, and replay for debugging.
\end{itemize}
\end{comment}

\section{Taxonomy of Agentic Memory}
\label{sec:anatomy}
We introduce a concise taxonomy of Memory-Augmented Generation organized around four memory structures: Lightweight Semantic, Entity-Centric and Personalized, Episodic and Reflective, and Structured and Hierarchical. Each category is further split into subcategories as shown in Figure~\ref{fig:mag-taxonomy} in Appendix~\ref{app:taxonomy}.

\subsection{Lightweight Semantic Memory}
Lightweight Semantic Memory is the simplest and most widely used form of MAG, where memory consists of independent textual units embedded in a vector space and retrieved via top-\textit{k} similarity search. Entries are typically append-only or minimally filtered, with no explicit structural relations between them.

\noindent \textbf{RL-Optimized Semantic Compression:} These schemes treat memory as a fixed-size semantic store and apply RL to optimize how information is compressed, retained, or overwritten under context constraints~\cite{wang2025mem, yan2025memory, yuan2025memsearcher}. Memory remains largely unstructured and textual, with learning focused on efficient content selection. For example, MemAgent~\cite{yu2025memagent} trains a latent token-level memory using multi-conversation RL to manage ultra-long contexts, while MemSearcher~\cite{yuan2025memsearcher} formulates multi-turn search and memory updates as an end-to-end RL problem, iteratively compressing semantic memory to enable scalable multi-hop reasoning without relying on full dialogue history.
\noindent \textbf{Heuristic / Prompt-Optimized:}
These approaches manage memory through prompt design or heuristic rewriting with a flat, compressed textual summary of prior steps generated via engineered instructions, reducing context length but remaining unstructured. Similar prompt-driven compression strategies appear in prior work~\cite{zhao2025ame,wu2025enhancing,liu2026simplemem,li2026hippocampusefficientscalablememory}. For example, ACON~\cite{kang2025acon} learns natural-language compression guidelines to selectively summarize long interaction histories, reducing context by up to 54\% without RL or fine-tuning, while CISM~\cite{liu2025compressed} condenses each reasoning and action step into compact semantic representations to enable long-horizon execution under context constraints without explicit external memory retrieval.

\noindent \textbf{Context Window Management:} This category manages the model’s working context within a single task, without accumulating memory across sessions. Prior interactions are folded, summarized, or reorganized to fit within a bounded window, prioritizing local reasoning efficiency over long-term storage or reuse~\cite{zhu2025llm,sakib2025memagent}. For example, AgentFold~\cite{ye2025agentfold} treats context as a dynamic workspace and learns multi-scale folding operations to condense long trajectories, while Context-Folding Agent~\cite{sun2025scaling} trains an RL-based policy that branches sub-tasks and compresses completed segments.
\noindent \textbf{Token-Level Semantic Memory:} This category encodes memory at the token level using dedicated memory tokens or compressed latent panels. These representations primarily capture semantic content, aiming to improve long-context handling with minimal overhead~\cite{wu2025tokmem, zhang2025memgen, yang2024text}. Memory entries are independent and inexpensive to store or retrieve, making them suitable for short- to medium-horizon recall, but limited for precise state tracking or long-term reasoning. For example, MemGen~\cite{zhang2025memgen} augments a frozen LLM with on-demand latent token memory via an RL-trained trigger and LoRA-based weaving, while TokMem~\cite{wu2025tokmem} replaces lengthy procedural prompts with trainable memory tokens to enable constant-size context management and scalable skill reuse.
\subsection{Entity-Centric and Personalized Memory}
Entity-centric and personalized memory organizes information around explicit entities such as users, tasks, or preferences, using structured records or attribute–value pairs. A predefined schema governs how information is stored, updated, and retrieved.

\noindent \textbf{Entity-Centric Memory:} Entity-centric memory organizes information around explicit entities and their attributes, maintaining structured, persistent records rather than raw dialogue~\cite{modarressi2023ret, liu2021rmm, liu2025webcoach}. For example, A-MEM~\cite{xu2025mem} builds interconnected knowledge notes with structured attributes and LLM-generated links; Memory-R1~\cite{yan2025memory} formulates entity memory management as an RL problem over a persistent entity–fact bank.%; and Mem0~\cite{chhikara2025mem0} performs entity-level fact extraction and consolidation, bridging entity-centric and graph-based memory designs.

\noindent \textbf{Personalized Memory:} Personalized memory maintains persistent user profiles that integrate short- and long-term preferences to support adaptive, identity-consistent behavior across sessions~\cite{zhong2024memorybank,li2025hello,kwon2025embodied,liu2025webcoach,mao2026bi,su2026beyond}. For example, PAMU~\cite{sun2025preference} combines sliding windows with moving averages to track evolving preferences, EgoMem~\cite{yao2025egomem} constructs lifelong multimodal profiles with conflict-aware updates, and MemOrb~\cite{huang2025memorb} stores compact reflective memories for continual improvement. %While effective for personalization, these systems remain sensitive to update errors and constrained by predefined schemas.

\subsection{Episodic and Reflective Memory}
Episodic and reflective memory adds temporal abstraction by organizing interactions into episodes or higher-level summaries. These systems periodically consolidate experience through summarization or reflection, producing compact representations of salient events over time.

\noindent \textbf{Episodic Buffer w/ Learned Control:} memory in these work consists of episodic interaction records maintained in a bounded buffer and dynamically inserted, retained, or deleted through learned policies~\cite{du2025memr, zhang2025memory, icarte2020act}. For example, MemR$^{3}$~\cite{du2025memr} models retrieval as a closed-loop retrieve–reflect–answer process; and the Act of Remembering~\cite{icarte2020act} formulates remembering as a control problem in POMDPs with a fixed-capacity episodic buffer. %; and MemAct~\cite{zhang2025memory} treats memory operations as RL actions over ID-addressable records, enabling autonomous working-memory curation for long-horizon tasks.

\noindent \textbf{Episodic Recall for Exploration:}
These methods leverage episodic memory to improve exploration and credit assignment in partially observable or long-horizon settings. Past experiences are stored and selectively retrieved to guide decision-making~\cite{na2024efficient, adamyan2025sam2rl}. For example, EMU~\cite{na2024efficient} maintains large-capacity episodic memories indexed by learned embeddings to accelerate cooperative MARL exploration, while SAM2RL~\cite{adamyan2025sam2rl} uses a visual memory bank as an episodic buffer and trains a PPO policy to manage memory replacement, outperforming heuristic updates under challenging conditions.

\noindent \textbf{Episodic Reflection \& Consolidation:} This subcategory reflects and consolidates episodic experiences into compact representations~\cite{tan2025prospect,kim2025pre,ouyang2025reasoningbank,dong2025towards,lee2024human}. The objective is to balance memory capacity with long-term reasoning utility. For example, MemP~\cite{fang2025memp} distills trajectories into procedural abstractions for continual refinement and transfer; LEGOMem~\cite{han2025legomem} constructs modular, role-aware procedural memories for multi-agent coordination; and TiMem~\cite{li2026timem} introduces a temporal-hierarchical memory tree for structured consolidation and scalable long-horizon personalization without RL or fine-tuning.

\noindent \textbf{Episodic Utility Learning:}
Episodic memories in these setting are augmented with learned value or utility signals that evolve over time, enabling selective retention and retrieval based on both semantic relevance and estimated long-term usefulness~\cite{zhou2025memento,cao2025remember}. For example, MemRL~\cite{zhang2026memrl} associates utility Q-values with intent–experience pairs and updates them online to balance stability and plasticity without fine-tuning, while Memory-T1~\cite{du2025memory} learns a temporal-aware retrieval policy via GRPO to optimize accuracy, grounding, and chronological consistency in long-context dialogue.

\subsection{Structured and Hierarchical Memory}
Structured and hierarchical memory systems impose explicit organization over stored information. Hierarchical designs partition memory into multiple tiers (e.g., short- and long-term stores), while structured approaches encode relationships among memory elements using graphs or other formal relational representations.

\noindent \textbf{Graph-Structured Memory:}
Graph-structured memory represents information as nodes and edges capturing semantic, temporal, causal, or entity-level relations, enabling reasoning over structured subgraphs~\cite{zhang2025g,zhang2025assomem,jiang2026synapse,tao2026membox,zhang2026implicit,hu2026memory}. This design supports multi-hop inference, provenance tracking, and coherent long-horizon reasoning. For example, MAGMA~\cite{jiang2026magma} organizes memory across semantic, temporal, causal, and entity graphs; Zep~\cite{rasmussen2025zep} constructs a bi-temporal knowledge graph with episodic and semantic layers; SGMem~\cite{wu2025sgmem} models dialogue as sentence-level graphs; and LatentGraphMem~\cite{zhang2026implicit} integrates latent graph encoding with a compact symbolic subgraph to balance stability, efficiency, and interpretability.

\noindent \textbf{OS-Inspired \& Hierarchical Memory:}
OS-inspired and hierarchical memory systems organize information into multi-tier storage layers (e.g., short-term, episodic, long-term), dynamically moving and consolidating data to balance scalability, retention, and adaptive forgetting~\cite{xu2025memory,ouyang2025can,zhang2025cogmem,jia2025pisa,li2026timem}. For example, MemGPT~\cite{packer2023memgpt} enables LLM-driven memory paging across tiers; MemoryOS~\cite{kang2025memory} implements a modular three-level hierarchy; EverMemOS~\cite{hu2026evermemos} and HiMem~\cite{zhang2026himem} consolidate episodic and semantic traces for long-horizon adaptation; and MeMAD~\cite{ling2025memad} stores structured debate experiences for reusable reasoning.

\noindent \textbf{Policy-Optimized Memory Management:}
Policy-optimized memory management treats storage, update, consolidation, and deletion as learnable decisions, using reinforcement learning or hybrid training to optimize long-horizon rewards~\cite{liu2025rcr,xu2025sedm,kang2025lm2,du2025memory}. For example, MEM1~\cite{zhou2025mem1} learns to maintain a compact internal state with constant-memory operations; and Mem-$\alpha$~\cite{wang2025mem} trains an RL policy to manage multi-component external memory under ultra-long contexts; and AtomMem~\cite{huo2026atommem} decomposes memory into CRUD actions to learn task-aligned control strategies. While enabling adaptive and scalable management, these approaches introduce greater system complexity and nontrivial maintenance overhead.

\subsection{Discussion}
The four categories described above capture the dominant memory structures used in contemporary MAG systems. While individual systems may combine multiple mechanisms, each can typically be characterized by a primary memory organization that governs its behavior. This structure-first taxonomy provides a foundation for understanding how design choices in agentic memory influence accuracy, efficiency, and reliability. In the next section, we build on this taxonomy to analyze the empirical limitations and pain points that arise across current MAG systems.
\section{Evaluation and Pain Points}
In this section, we move beyond taxonomy to empirically analyze the practical bottlenecks hindering robust deployment. While theoretical architectures are promising, real world utility is strictly constrained by evaluation validity, system efficiency, and backbone reliability. We dissect these challenges across four critical dimensions:
\begin{enumerate}[leftmargin=*, itemsep=0pt, topsep=0pt]
    \item Benchmark Validity: Are we testing memory or just context length?
    \item Metric Reliability: Can lexical metrics capture semantic coherence?
    \item System Efficiency: The ``Agency Tax'' of latency and cost.
    \item Backbone Sensitivity: The ``Silent Failure'' of memory operations in open-weight models.
\end{enumerate}

\subsection{Experimental Setup}
\label{subsec:setup}
We evaluate representative MAG systems spanning the four taxonomy categories introduced in Section~\ref{sec:anatomy}. Six memory architectures are selected: AMem~\cite{xu2025mem}, MemoryOS~\cite{kang2025memory}, Nemori~\cite{nan2025nemori}, MAGMA~\cite{jiang2026magma}, SimpleMEM~\cite{liu2026simplemem} and MemSkill~\cite{zhang2026memskill} as shown in Table~\ref{tab:baselines} of Appendix~\ref{app:baselines}. All systems are configured to follow their default or recommended settings, except where modifications are required to ensure comparability. We employ a suite of Large Language Models (LLMs) to serve as the agent controller including \texttt{gpt-4o-mini}~\cite{achiam2023gpt} and \texttt{Qwen-2.5-3B}~\cite{yang2024qwen2}.

\label{app:baselines}

\subsection{Benchmark Scalability: The Context Saturation Risk}
\label{subsec:benchmark_gap}
\begin{table*}[t]
\centering
\small
\caption{Structural saturation risk of memory benchmarks. Benchmarks are analyzed based on intrinsic statistics rather than model performance. \textit{Saturation Risk} is a heuristic estimate of whether a long-context LLM may solve the benchmark through direct prompting.}
\label{tab:benchmark_characteristics}
\resizebox{\textwidth}{!}{
\begin{tabular}{l | c c c | c}
\toprule
\multirow{2}{*}{\textbf{Benchmark}} & \multicolumn{3}{c|}{\textbf{Scalability Dimensions}} & \multirow{2}{*}{\textbf{Theoretical Saturation Risk}} \\
 & \textbf{Avg. Volume} & \textbf{Interaction Depth} & \textbf{Entity Diversity} & \\
\midrule

\textbf{HotpotQA~\cite{yang2018hotpotqa}} 
& $\sim$1k Tokens & Single Turn & Low 
& High (Trivial for Context Window) \\

\textbf{LoCoMo~\cite{maharana2024evaluating}} 
& $\sim$20k Tokens & 35 Sessions & High 
& Moderate (Requires Reasoning) \\

\textbf{LongMemEval-S~\cite{wu2024longmemeval}} 
& 103k Tokens & 5 Core Abilities & High 
& Moderate (Borderline) \\

\textbf{LongMemEval-M~\cite{wu2024longmemeval}} 
& $>$1M Tokens & 5 Core Abilities & High 
& Low (Requires External Memory) \\

\textbf{MemBench~\cite{tan2025membench}} 
& $\sim$100k Tokens & Fact/Reflection & Medium 
& High (Fits in 128k Window) \\

\bottomrule
\end{tabular}
}
\end{table*}

A key motivation for agentic memory is to support reasoning beyond a model’s finite context window. Yet as LLM windows expand (e.g., 128k to 1M tokens), many benchmarks risk \textit{context saturation}, where all relevant information fits within a single prompt, making external memory seemingly unnecessary. In this section, rather than comparing performance, we examine the intrinsic properties of existing datasets to evaluate their continued validity in the long-context era.

\subsubsection{Dimensions of Limitation}
First, we evaluate benchmark scalability along three structural axes: volume, interaction depth, and entity diversity, to assess their saturation risk under long-context LLMs as shown in Table~\ref{tab:benchmark_characteristics}.

\textbf{Volume (Total Token Load).} This dimension captures the aggregate information size a model must process. Benchmarks such as \textit{HotpotQA} ($\sim$1k tokens) and \textit{MemBench} ($\sim$100k tokens) fall within a 128k context window, implying high theoretical saturation risk. \textit{LoCoMo} ($\sim$20k tokens) similarly remains comfortably in-window for modern models. Only datasets that substantially exceed the active window (e.g., \textit{LongMemEval-M} at >1M tokens) structurally require external memory.

\textbf{Interaction Depth (Temporal Structure).} Beyond raw volume, scalability depends on how information unfolds across sessions. Single-turn QA (e.g., \textit{HotpotQA}) imposes minimal temporal dependency, whereas multi-session settings (e.g., \textit{LoCoMo} with 35 sessions) introduce longitudinal reasoning. However, unless cross-session dependencies exceed the active window or require persistent state tracking beyond prompt capacity, such datasets may still be solvable through direct in-context aggregation rather than true memory management.

\textbf{Entity Diversity (Relational Complexity).} This axis measures how many distinct entities or conceptual threads must be tracked simultaneously. Low-diversity benchmarks permit near-isolated retrieval, while higher-diversity settings (e.g., \textit{LoCoMo}, \textit{LongMemEval}) increase interference and relational reasoning demands. Nevertheless, if entity interactions remain bounded within context limits, structured external memory may not be strictly necessary.

\textbf{Discussion:} Taken together, these dimensions show that saturation risk is determined not by surface difficulty but by whether a benchmark’s structural properties exceed the representational capacity of long-context LLMs.

\subsubsection{Context Saturation Gap as an Empirical Diagnostic}
To address these limitations, we propose that future evaluations explicitly quantify the \textit{Context Saturation Gap} ($\Delta$), defined as the performance difference between a Memory-Augmented Agent (MAG) and a brute-force Full-Context baseline under the same backbone and evaluation protocol:
\begin{equation}
    \Delta = \text{Score}_{\text{MAG}} - \text{Score}_{\text{FullContext}}
\end{equation}

A large positive $\Delta$ indicates that external memory provides an advantage beyond simply placing all available evidence in the prompt, especially in out-of-window or lost-in-the-middle regimes. However, $\Delta$ should be viewed as a diagnostic rather than a strict pass/fail criterion, since benchmarks may still evaluate memory through efficiency, updateability, robustness, or evidence faithfulness even when Full-Context remains competitive.

Table~\ref{tab:benchmark_characteristics} therefore reports structural saturation risk rather than an empirical saturation test. Datasets with limited volume and shallow complexity should be paired with Full-Context baselines before drawing strong conclusions about memory-specific benefits.
\subsection{LLM-as-a-Judge Evaluation}

\begin{table*}[t]
\centering
\small
\caption{Robustness of system ranking across evaluation protocols. We compare Lexical metrics (F1) against LLM-based semantic evaluation using three distinct prompt sources: MAGMA, Nemori, and SimpleMem.}% \textbf{Key Insight:} While absolute scores vary, MAGMA consistently ranks 1st or a close 2nd across diverse grading criteria. In contrast, lexical metrics (F1) produce misleading rankings, such as underrating the semantically capable AMem.}
\label{tab:metric_robustness}
\resizebox{\textwidth}{!}{
\begin{tabular}{l | c c | c c c }
\toprule
\multirow{2}{*}{\textbf{Method}} & \multicolumn{2}{c|}{\textbf{Lexical Metric}} & \multicolumn{3}{c}{\textbf{Semantic Judge Score (Rank)}} \\
 & \textbf{F1-Score} & \textbf{Rank} & \textbf{Prompt 1 (MAGMA)} & \textbf{Prompt 2 (Nemori)} & \textbf{Prompt 3 (SimpleMem)} \\
\midrule
AMem~\cite{xu2025mem}          & 0.116 & 5 & 0.480 (4) & 0.512 (4) & 0.482 (4) \\
MemoryOS~\cite{kang2025memory}       & 0.413 & 3 & 0.553 (3) & 0.589 (3) & 0.552 (3) \\
Nemori~\cite{nan2025nemori}        & \textbf{0.502} & \textbf{1} & 0.602 (2) & \textbf{0.781 (1)} & 0.649 (2) \\
MAGMA~\cite{jiang2026magma}  & 0.467 & 2 & \textbf{0.670 (1)} & 0.741 (2) & \textbf{0.665 (1)} \\
SimpleMEM~\cite{liu2026simplemem}      & 0.268 & 4 & 0.294 (5) & 0.298 (5) & 0.289 (5) \\
MemSkill~\cite{zhang2026memskill}
& 0.082 & 6
& 0.221 (6) & 0.220 (6) & 0.127 (6) \\
\bottomrule
\end{tabular}
}
\vspace{-10pt}
\end{table*}

Traditional lexical metrics (e.g., F1, BLEU) emphasize surface-level token overlap, which is insufficient for agentic memory tasks where the goal is accurate retrieval and coherent synthesis rather than exact phrasing. To better capture semantic correctness, we adopt an LLM-based evaluator (gpt-4o-mini) as a proxy for human judgment. In this section, we assess the reliability of this protocol by analyzing the misalignment between lexical and semantic metrics and demonstrating the stability of our system rankings across competitive evaluation settings.

\subsubsection{The Misalignment Gap}
Do lexical metrics correctly identify the best memory system? To examine this, we compared system rankings produced by F1-score with those generated by an LLM-based judge across six representative architectures on the \textit{LoCoMo} dataset.

Table~\ref{tab:metric_robustness} (Left) reveals a significant disconnect. Lexical metrics often fail to capture the strengths of abstractive memory systems. For example, AMem achieves solid semantic performance (Rank 4 across prompts) due to its logical coherence, yet it is heavily penalized by F1 (Rank 5, Score 0.116) because it does not rely on verbatim overlap. In contrast, SimpleMem receives a relatively higher F1 score (0.268) despite demonstrating limited ability to synthesize complex answers (semantic score < 0.30). This divergence indicates that optimizing solely for F1 may favor surface-level memorization over genuine reasoning and memory integration.

\subsubsection{Semantic Judge Robustness Across Prompts}
\label{subsubsec:promptsdiff}
A common concern with LLM-as-a-judge is “prompt overfitting,” where a system appears strong only under a specific grading instruction. To ensure fairness and generality, we evaluated all architectures using three distinct prompt protocols derived from different sources (details in Appendix~\ref{sec:LLM-as-a-Judge}).

As shown in Table~\ref{tab:metric_robustness} (Right), compared with F1-based rankings, the semantic judge exhibits strong robustness: the relative ordering of architectures remains highly consistent across different rubrics. While absolute scores fluctuate due to variations in grading strictness and prompt formulation, the comparative conclusions remain stable.

\subsubsection{Discussion}
Lexical metrics provide a convenient baseline but systematically diverge from semantic judgments due to two core failure modes: the Paraphrase Penalty, where correct abstractive answers are penalized for low token overlap, and the Negation Trap, where high overlap masks factual errors. Detailed examples are provided in Appendix~\ref{app:metric_cases}.

In contrast, the semantic judge demonstrates greater stability: architecture rankings remain consistent across different grading rubrics, suggesting it better reflects underlying memory quality rather than surface phrasing. Although absolute scores vary with prompt strictness and some models show rubric-aligned specialization, the relative ordering is robust.

Overall, these results support LLM-as-a-judge as a more reliable evaluation protocol for agentic memory, while highlighting the importance of careful prompt design.

\subsection{Backbone Sensitivity and Format Stability}
\label{subsec:backbone}
Agentic memory requires the backbone model to both answer queries and execute structured memory operations (e.g., updates and consolidation). Long-term stability thus depends on reliable adherence to strict output formats. To evaluate this “Stability Gap,” we compare representative memory architectures using an API model (\texttt{gpt-4o-mini}) and an open-weight model (\texttt{Qwen-2.5-3B}).

Table~\ref{tab:backbone_sensitivity} reveals a clear divergence driven by invalid structured outputs (e.g., malformed JSON, hallucinated keys) during memory maintenance:  \textbf{1) Instruction Following vs. Reasoning:} While \texttt{Qwen-2.5-3B} demonstrates basic capability in conversational reasoning, it experiences a noticeable drop in End-Task Answer Scores and exhibits a significantly higher format error rate during memory updates compared to \texttt{gpt-4o-mini}. This ``Silent Failure'' implies that while the agent can converse fluently in the short term, its long-term memory becomes corrupted due to failed write operations. \textbf{2) Method Sensitivity:} The impact of the backbone varies by architecture complexity. Append-only systems are relatively robust, as they require minimal structured generation. In contrast, graph-based and episodic architectures are highly sensitive: extracting entities, constructing relations, and performing logical deduplication significantly increase format errors under weaker backbones, often leading to structural instability or collapse in memory maintenance.

\begin{table}[t]
\centering
\small
\caption{Backbone Sensitivity Analysis. Frequency of recoverable format deviations during memory operations is used. Higher values indicate greater reliance on fallback parsing due to inconsistent structured outputs.}
\label{tab:backbone_sensitivity}
\resizebox{\columnwidth}{!}{
\begin{tabular}{l l | c c}
\toprule
\textbf{Backbone} & \textbf{Method} 
& \textbf{Answer Score} 
& \textbf{ Format Error}  \\
\midrule
\multirow{2}{*}{\textbf{gpt-4o-mini}} 
 & SimpleMem &0.289  & 1.20\% \\
 & Nemori &0.781  &  17.91\%\\
\midrule
\multirow{2}{*}{\textbf{Qwen-2.5-3B}} 
 & SimpleMem &0.102  & 4.82\% \\
 & Nemori & 0.447 &  30.38\%\\
\bottomrule
\end{tabular}
}
\vspace{-10pt}
\end{table}

\subsection{System Performance Evaluation}
\label{subsec:system_perf}

\begin{table*}[t]
\centering
\small
\caption{\textbf{The "Agency Tax": Efficiency Profiling.} We evaluate the trade-off between runtime user latency and offline construction cost. User Latency($T_{read} + T_{gen}$) dictates the interactive experience, while Construction Cost reflects the scalability and economic feasibility of the system. Note that Maintenance Cost is omitted as it is often handled asynchronously.}
\label{tab:latency_breakdown}
\setlength{\tabcolsep}{8pt} 
\begin{tabular*}{\textwidth}{@{\extracolsep{\fill}} l c c c | c c @{}}
\toprule
\multirow{2}{*}{\textbf{Method}} &
\multicolumn{3}{c|}{\textbf{User-Facing Latency (per turn)}} &
\multicolumn{2}{c}{\textbf{Construction Cost (Offline)}} \\
\cmidrule(lr){2-4} \cmidrule(l){5-6}
 & \textbf{Retrieval ($T_{read}$)} & \textbf{Generation ($T_{gen}$)} & \textbf{Total (s)} & \textbf{Time (h)} & \textbf{Tokens (k)} \\
\midrule

Full Context
& N/A & 1.726 & 1.726
& N/A & N/A \\

LOCOMO~\cite{maharana2024evaluating}
& 0.415 & 0.368 & 0.783
& 0.86 & 1,623 \\

AMem~\cite{xu2025mem}
& 0.062 & 1.119 & 1.181
& 15.00 & 1,486 \\

MemoryOS~\cite{kang2025memory}
& 31.247 & 1.125 & 32.372
& 7.83 & 4,043 \\

Nemori~\cite{nan2025nemori}
& 0.254 & 0.875 & 1.129 
& 3.25 & 7,044 \\

MAGMA~\cite{jiang2026magma}
& 0.497 & 0.965 & 1.462
& 7.28 & 2,725 \\

SimpleMem~\cite{liu2026simplemem}
& 0.009 & 1.048 & 1.057
& 3.45 & 1,308 \\

MemSkill~\cite{zhang2026memskill}
& 0.005 & 0.301 & 0.306
& 0.60 & 1,796 \\
\bottomrule
\end{tabular*}
\vspace{-5pt}
\end{table*}
While accuracy is critical, the practical viability of agentic memory is constrained by latency and cost. Unlike read-only RAG systems, agentic memory introduces maintenance operations such as memory extraction, updates, and consolidation. We measure the user-facing load as retrieval ($T_{read}$), covering search and traversal, and generation ($T_{gen}$), including context processing and token decoding, while discussing maintenance overhead qualitatively.

In this section, we quantify user-facing latency ($T_{read} + T_{gen}$) and offline scalability using Table~\ref{tab:latency_breakdown}, while highlighting maintenance as a hidden system-level bottleneck.

\subsubsection{Latency and Maintenance Trade-offs in MAG}
We analyze the end-to-end user-perceived latency ($T_{read} + T_{gen}$) alongside the often-overlooked maintenance overhead ($T_{write}$). Although Full Context eliminates retrieval cost, it incurs the highest generation latency ($T_{gen} \approx 1.73$s), confirming that large pre-fill computation increases time-to-first-token. Lightweight systems such as SimpleMem and LOCOMO achieve sub-second latency ($<1.1$s) through efficient indexing, while MAGMA maintains a balanced profile ($\sim1.46$s), adding modest overhead for graph traversal. In contrast, MemoryOS emerges as a clear bottleneck, with latency exceeding 32 seconds, suggesting that strict hierarchical paging (e.g., STM→LTM recursion) is impractical for interactive settings.

Beyond user-facing latency, maintenance introduces a hidden scalability constraint. Append-only or lightweight systems incur lower update cost, whereas structured architectures (e.g., MAGMA, AMem) require graph restructuring and LLM-driven consolidation. Although often asynchronous, excessive maintenance overhead risks throughput collapse, where updates lag behind user interactions and memory becomes stale. Thus, structured memory demands robust asynchronous infrastructure to remain viable at scale.

\subsubsection{Offline Scalability: Time and Token Economics}
Beyond online latency, we evaluate the offline cost of building the memory index. AMem requires approximately 15 hours for construction far slower than other baselines, suggesting super-linear update complexity (e.g., pairwise consolidation) that limits scalability on large datasets.

Token consumption further exposes cost trade-offs. Nemori uses over 7.04M tokens during index construction, nearly five times that of SimpleMem (1.3M). Although this yields strong accuracy, it reflects a substantial “intelligence tax,” where improved memory quality incurs significantly higher operational cost. In comparison, MAGMA achieves a more favorable Pareto balance, delivering robust performance with moderate token usage (2.7M).

% =================================================
% ==================================================

\section{Conclusion and Future Directions}
\label{sec:conclusion}
Our analysis shows that agentic memory is limited not only by architecture, but also by evaluation validity, scalability, and robustness.

\paragraph{1. Rethinking Benchmark and Evaluation Design.}
Future benchmarks should be saturation-aware. As context windows expand, full-context prompting may solve many tasks, making it harder to isolate the value of external memory. Benchmarks should therefore stress task volume, temporal depth, entity diversity, and long-range dependency, while using the Context Saturation Gap ($\Delta$) as a diagnostic signal.

Evaluation should also move beyond lexical overlap. F1-style metrics often miss semantic correctness, while LLM-as-a-judge requires prompt calibration and robustness checks.

\paragraph{2. Designing Scalable and Robust Agentic Memory Systems.}
Agentic memory systems must balance accuracy, latency, cost, and reliability. Structured memory improves reasoning but introduces maintenance overhead, while lightweight approaches improve efficiency but may lack abstraction.

Future systems should explicitly model write latency, maintenance throughput, and user-facing cost. They should also use backbone-aware memory operations, constrained decoding, validation layers, or adaptive schemas to reduce silent corruption.

\section*{Limitations}
This survey has several limitations. First, agentic memory is evolving rapidly, so our taxonomy may miss concurrent or very recent systems. Second, our empirical analysis covers representative MAG architectures and selected benchmarks rather than all systems and settings; results may vary with implementations, prompts, model versions, and API behavior. These limitations call for broader, standardized evaluations of agentic memory systems.

% ==================================================
% References (ACL style)
% ==================================================
\bibliography{custom}

% ==================================================
% Appendices
% ==================================================
\clearpage
\appendix

\section{Taxonomy of Agentic Memory}
\label{app:taxonomy}

Figure~\ref{fig:mag-taxonomy} provides a comprehensive visual taxonomy of recent advancements in Memory-Augmented Generation (MAG). Given the rapid proliferation of memory architectures for LLM agents, this tree diagram categorizes contemporary literature into four distinct structural paradigms:

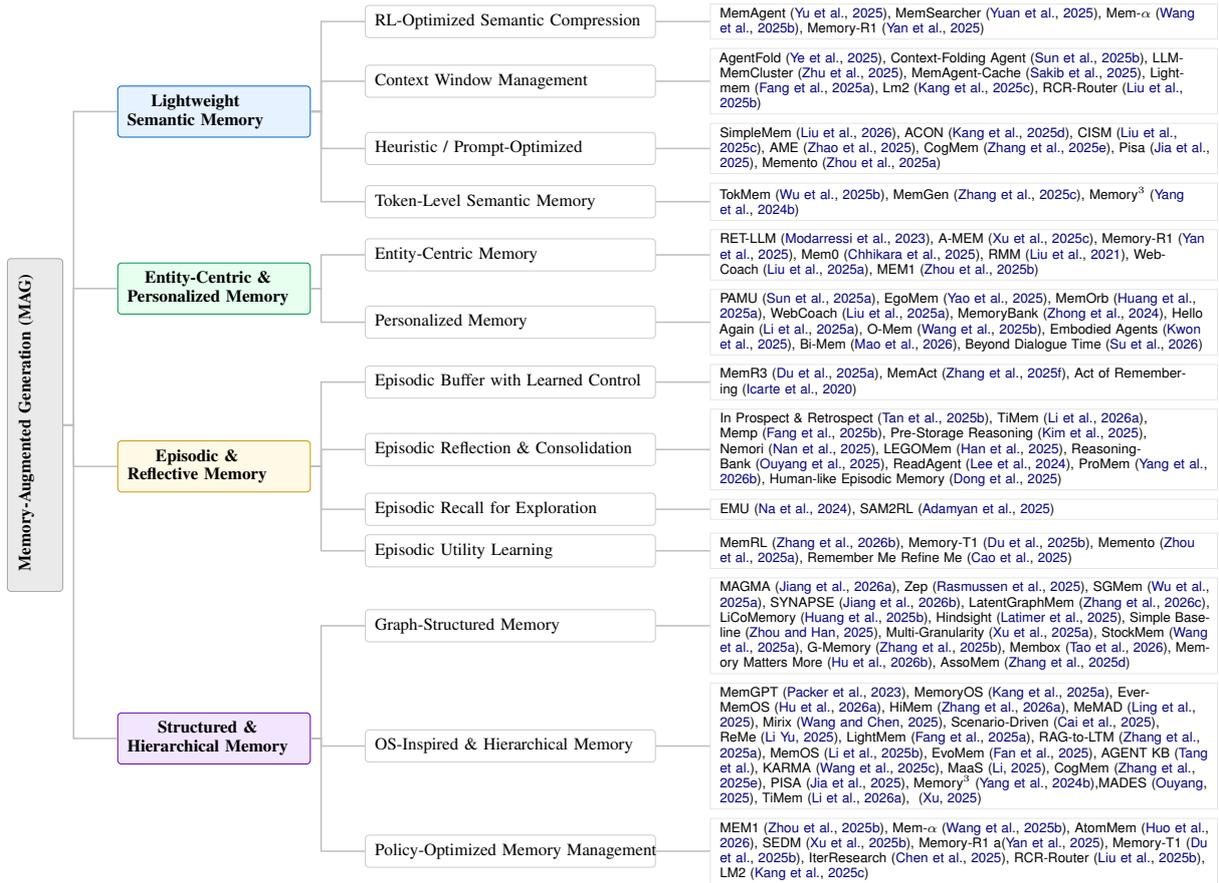
\begin{figure*}[!t]
\centering
\resizebox{\textwidth}{!}{%
    \begin{forest}
      for tree={
        grow=east,
        reversed=true,
        parent anchor=east,
        child anchor=west,
        anchor=west,
        edge={-, thick, draw=gray!50},
        l sep=10mm,
        s sep=1.6mm,
        inner xsep=5pt,
        inner ysep=3.5pt,
        font=\small,
      },
      forked edges,
      % ROOT
      [{\rotatebox{90}{\textbf{Memory-Augmented Generation (MAG)}}},
        draw=gray!70, fill=rootfill, rounded corners=2pt,
        text width=6mm, align=center, font=\small\bfseries,
        inner xsep=6pt, inner ysep=8pt,
        %
        % ===== 3.1 (BLUE) =====
        [{\textbf{Lightweight}\\\textbf{Semantic Memory}},
          draw=l1blue, fill=l1bluefill, rounded corners=2pt,
          text width=32mm, align=center, font=\small\bfseries,
          [{RL-Optimized Semantic Compression},
            draw=gray!40, fill=white, rounded corners=2pt,
            text width=50mm, align=center,
            [{{\scriptsize\sffamily MemAgent~\cite{yu2025memagent}, MemSearcher~\cite{yuan2025memsearcher}, Mem-$\alpha$~\cite{wang2025mem}, Memory-R1~\cite{yan2025memory}}},
              draw=gray!20, fill=white, text width=90mm, font=\scriptsize\sffamily, inner ysep=2pt]
          ]
          [{Context Window Management},
            draw=gray!40, fill=white, rounded corners=2pt,
            text width=50mm, align=center,
            [{{\scriptsize\sffamily AgentFold~\cite{ye2025agentfold}, Context-Folding Agent~\cite{sun2025scaling}, LLM-MemCluster~\cite{zhu2025llm}, MemAgent-Cache~\cite{sakib2025memagent}, Lightmem~\cite{fang2025lightmem}, Lm2~\cite{kang2025lm2}, RCR-Router~\cite{liu2025rcr}}},
              draw=gray!20, fill=white, text width=90mm, font=\scriptsize\sffamily, inner ysep=2pt]
          ]
          [{Heuristic / Prompt-Optimized},
            draw=gray!40, fill=white, rounded corners=2pt,
            text width=50mm, align=center,
            [{{\scriptsize\sffamily SimpleMem~\cite{liu2026simplemem}, ACON~\cite{kang2025acon}, CISM~\cite{liu2025compressed}, AME~\cite{zhao2025ame}, CogMem~\cite{zhang2025cogmem}, Pisa~\cite{jia2025pisa}, Memento~\cite{zhou2025memento}}},
              draw=gray!20, fill=white, text width=90mm, font=\scriptsize\sffamily, inner ysep=2pt]
          ]
          [{Token-Level Semantic Memory},
            draw=gray!40, fill=white, rounded corners=2pt,
            text width=50mm, align=center,
            [{{\scriptsize\sffamily TokMem~\cite{wu2025tokmem}, MemGen~\cite{zhang2025memgen}, $\text{Memory}^3$~\cite{yang2024text}}},
              draw=gray!20, fill=white, text width=90mm, font=\scriptsize\sffamily, inner ysep=2pt]
          ]
        ]
        %
        % ===== 3.2 (GREEN) =====
        [{\textbf{Entity-Centric \&}\\\textbf{Personalized Memory}},
          draw=l1green, fill=l1greenfill, rounded corners=2pt,
          text width=32mm, align=center, font=\small\bfseries,
          [{Entity-Centric Memory},
            draw=gray!40, fill=white, rounded corners=2pt,
            text width=50mm, align=center,
            [{{\scriptsize\sffamily RET-LLM~\cite{modarressi2023ret}, A-MEM~\cite{xu2025mem}, Memory-R1~\cite{yan2025memory}, Mem0~\cite{chhikara2025mem0}, RMM~\cite{liu2021rmm}, WebCoach~\cite{liu2025webcoach}, MEM1~\cite{zhou2025mem1}}},
              draw=gray!20, fill=white, text width=90mm, font=\scriptsize\sffamily, inner ysep=2pt]
          ]
          [{Personalized Memory},
            draw=gray!40, fill=white, rounded corners=2pt,
            text width=50mm, align=center,
            [{{\scriptsize\sffamily PAMU~\cite{sun2025preference}, EgoMem~\cite{yao2025egomem}, MemOrb~\cite{huang2025memorb}, WebCoach~\cite{liu2025webcoach}, MemoryBank~\cite{zhong2024memorybank}, Hello Again~\cite{li2025hello}, O-Mem~\cite{wang2025mem}, Embodied Agents~\cite{kwon2025embodied}, Bi-Mem~\cite{mao2026bi}, Beyond Dialogue Time~\cite{su2026beyond}}},
              draw=gray!20, fill=white, text width=90mm, font=\scriptsize\sffamily, inner ysep=2pt]
          ]
        ]
        %
        % ===== 3.3 (YELLOW) =====
        [{\textbf{Episodic \&}\\\textbf{Reflective Memory}},
          draw=l1yellow, fill=l1yellowfill, rounded corners=2pt,
          text width=32mm, align=center, font=\small\bfseries,
          [{Episodic Buffer with Learned Control},
            draw=gray!40, fill=white, rounded corners=2pt,
            text width=50mm, align=center,
            [{{\scriptsize\sffamily MemR3~\cite{du2025memr}, MemAct~\cite{zhang2025memory}, Act of Remembering~\cite{icarte2020act}}},
              draw=gray!20, fill=white, text width=90mm, font=\scriptsize\sffamily, inner ysep=2pt]
          ]
          [{Episodic Reflection \& Consolidation},
            draw=gray!40, fill=white, rounded corners=2pt,
            text width=50mm, align=center,
            [{{\scriptsize\sffamily In Prospect \& Retrospect~\cite{tan2025prospect}, TiMem~\cite{li2026timem}, Memp~\cite{fang2025memp}, Pre-Storage Reasoning~\cite{kim2025pre}, Nemori~\cite{nan2025nemori}, LEGOMem~\cite{han2025legomem}, ReasoningBank~\cite{ouyang2025reasoningbank}, ReadAgent~\cite{lee2024human}, ProMem~\cite{yang2026beyond}, Human-like Episodic Memory~\cite{dong2025towards},
            MemSkill~\cite{zhang2026memskill}}},
              draw=gray!20, fill=white, text width=90mm, font=\scriptsize\sffamily, inner ysep=2pt]
          ]
          [{Episodic Recall for Exploration},
            draw=gray!40, fill=white, rounded corners=2pt,
            text width=50mm, align=center,
            [{{\scriptsize\sffamily EMU~\cite{na2024efficient}, SAM2RL~\cite{adamyan2025sam2rl}}},
              draw=gray!20, fill=white, text width=90mm, font=\scriptsize\sffamily, inner ysep=2pt]
          ]
          [{Episodic Utility Learning},
            draw=gray!40, fill=white, rounded corners=2pt,
            text width=50mm, align=center,
            [{{\scriptsize\sffamily MemRL~\cite{zhang2026memrl}, Memory-T1~\cite{du2025memory}, Memento~\cite{zhou2025memento}, Remember Me Refine Me~\cite{cao2025remember}}},
              draw=gray!20, fill=white, text width=90mm, font=\scriptsize\sffamily, inner ysep=2pt]
          ]
        ]
        %
        % ===== 3.4 (PURPLE) =====
        [{\textbf{Structured \&}\\\textbf{Hierarchical Memory}},
          draw=l1purple, fill=l1purplefill, rounded corners=2pt,
          text width=32mm, align=center, font=\small\bfseries,
          [{Graph-Structured Memory},
            draw=gray!40, fill=white, rounded corners=2pt,
            text width=50mm, align=center,
            [{{\scriptsize\sffamily MAGMA~\cite{jiang2026magma}, Zep~\cite{rasmussen2025zep}, SGMem~\cite{wu2025sgmem}, SYNAPSE~\cite{jiang2026synapse}, LatentGraphMem~\cite{zhang2026implicit}, LiCoMemory~\cite{huang2025licomemory}, Hindsight~\cite{latimer2025hindsight}, Simple Baseline~\cite{zhou2025simple}, Multi-Granularity~\cite{xu2025single}, StockMem~\cite{wang2025stockmem}, G-Memory~\cite{zhang2025g}, Membox~\cite{tao2026membox}, Memory Matters More~\cite{hu2026memory}, AssoMem~\cite{zhang2025assomem},
            HAGE~\cite{jiang2026hage}}},
              draw=gray!20, fill=white, text width=90mm, font=\scriptsize\sffamily, inner ysep=2pt]
          ]
          [{OS-Inspired \& Hierarchical Memory},
            draw=gray!40, fill=white, rounded corners=2pt,
            text width=50mm, align=center,
            [{{\scriptsize\sffamily MemGPT~\cite{packer2023memgpt}, MemoryOS~\cite{kang2025memory}, EverMemOS~\cite{hu2026evermemos}, HiMem~\cite{zhang2026himem}, MeMAD~\cite{ling2025memad}, Mirix~\cite{wang2025mirix}, Scenario-Driven~\cite{cai2025scenario}, ReMe~\cite{AgentscopeReMe2025}, LightMem~\cite{fang2025lightmem}, RAG-to-LTM~\cite{zhang2025conversational}, MemOS~\cite{li2025memos}, EvoMem~\cite{fan2025evomem}, AGENT~KB~\cite{tang2025agent}, KARMA~\cite{wang2025karma}, MaaS~\cite{li2025memory}, CogMem~\cite{zhang2025cogmem}, PISA~\cite{jia2025pisa}, Memory$^3$~\cite{yang2024text},MADES~\cite{ouyang2025can}, TiMem~\cite{li2026timem},
            ~\cite{xu2025memory}}},
              draw=gray!20, fill=white, text width=90mm, font=\scriptsize\sffamily, inner ysep=2pt]
          ]
          [{Policy-Optimized Memory Management},
            draw=gray!40, fill=white, rounded corners=2pt,
            text width=50mm, align=center,
            [{{\scriptsize\sffamily 
            MEM1~\cite{zhou2025mem1}, Mem-$\alpha$~\cite{wang2025mem}, AtomMem~\cite{huo2026atommem}, SEDM~\cite{xu2025sedm}, 
            Memory-R1~a\cite{yan2025memory},
            Memory-T1~\cite{du2025memory}, IterResearch~\cite{chen2025iterresearch}, RCR-Router~\cite{liu2025rcr}, LM2~\cite{kang2025lm2},
            MemSkill~\cite{zhang2026memskill},
            HAGE~\cite{jiang2026hage}
            }},
              draw=gray!20, fill=white, text width=90mm, font=\scriptsize\sffamily, inner ysep=2pt]
          ]
        ]
      ]
    \end{forest}
}
\caption{Taxonomy of Memory-Augmented Generation (MAG) systems.}
\label{fig:mag-taxonomy}
\end{figure*}

\section{Agentic Memory Background}\label{sec:basic_behaviors}

\subsection{Memory Operations in Agentic Systems}
We characterize \emph{agentic memory} as an external, non-parametric subsystem that interacts with an agent through two coupled processes: \textbf{inference-time recall} (reading memory to condition decisions) and \textbf{memory update} (writing, consolidating, and forgetting to maintain a useful long-term store). Let $f_\theta$ denote a frozen foundation model (or policy model) with parameters $\theta$, and let $\mathcal{M}_t$ denote the external memory state at step $t$. Given an observation $o_t$ (e.g., user input, tool output, sensor data) and agent state $s_t$ (e.g., goals, plans, tool traces), the agent first produces a query $q_t$ and recalls relevant memory:
\begin{align}
q_t &= \mathrm{Query}(o_t, s_t), \label{eq:query_basic}\\
r_t &= \mathrm{Read}(\mathcal{M}_t, q_t). \label{eq:read_basic}
\end{align}
The retrieved content $r_t$ is then integrated into the model input to produce an action or response:
\begin{equation}
a_t \sim \pi_\theta(o_t, r_t, s_t)
\;=\;
f_\theta\!\Big(\phi(o_t, s_t)\ \oplus\ \psi(r_t)\Big),
\label{eq:policy_basic}
\end{equation}
where $\phi(\cdot)$ formats current context, $\psi(\cdot)$ formats retrieved memory, and $\oplus$ denotes an integration operator (e.g., concatenation, schema-based slots, or cross-modal fusion). This abstraction makes explicit that agentic memory influences behavior through an external recall term rather than by updating $\theta$.

\paragraph{Inference-time retrieval as approximate utility optimization.}
External memory retrieval typically selects items $\{m_i\}_{i=1}^N$ from $\mathcal{M}_t$ using a scoring function (dense similarity, sparse matching, or reranking). A common instantiation is top-$k$ recall:
\begin{equation}
r_t
=
\mathrm{TopK}\Big(\{m_i\}_{i=1}^N;\ \mathrm{score}(q_t,m_i),\ k\Big).
\label{eq:topk_basic}
\end{equation}
However, in agentic settings, the ideal ``relevance'' is not purely semantic but \emph{decision-conditional}. One can express an idealized retrieval objective as selecting memory that maximizes downstream utility:
\begin{equation}
r_t^\star
=
\arg\max_{r \subseteq \mathcal{M}_t}
\ \mathbb{E}\big[\,U(a_t \mid o_t, r, s_t)\,\big],
\label{eq:utility_retrieval_basic}
\end{equation}
where $U(\cdot)$ denotes agent utility (e.g., task success, efficiency, robustness). Practical systems approximate \eqref{eq:utility_retrieval_basic} using similarity search, learned rerankers, multi-hop retrieval, planner-guided recall, or retrieval policies trained to better align $\mathrm{score}(\cdot)$ with utility.

\paragraph{Memory update as explicit memory actions.}
After producing an action $a_t$ (and possibly observing its outcome), the agent updates external memory through a write function:
\begin{equation}
\mathcal{M}_{t+1}=\mathrm{Write}(\mathcal{M}_t, o_t, a_t, s_t).
\label{eq:write_basic}
\end{equation}
It is often useful to make updates explicit as \emph{memory actions}. Let $u_t \in \mathcal{U}$ denote a memory action such as \textsc{store}, \textsc{update}, \textsc{summarize}, \textsc{link}, \textsc{evict}, or \textsc{delete}. Then:
\begin{equation}
u_t = g(o_t, a_t, s_t), \qquad \mathcal{M}_{t+1}=T(\mathcal{M}_t, u_t),
\label{eq:memory_action_basic}
\end{equation}
where $g(\cdot)$ may be rule-based, model-driven, or learned, and $T$ applies the chosen action to the memory store. This view connects naturally to RL-guided memory, where $g(\cdot)$ can be optimized as a policy over memory actions.
\section{Related Work}
\label{app:relatedwork}
Several recent surveys have examined memory mechanisms for Agentic AI systems, each from a distinct vantage point. The AI Hippocampus~\cite{jia2026ai} presents a broad synthesis organized around a brain-inspired trichotomy of implicit memory, explicit memory, and agentic memory, further extending the analysis to multimodal settings involving vision, audio, and embodied interaction.
Memory in the Age of AI Agents~\cite{hu2025memory} proposes a ``forms--functions--dynamics'' framework that
categorizes agent memory along three orthogonal axes: architectural form, functional role, and lifecycle dynamics, providing a comprehensive conceptual vocabulary for the
rapidly fragmenting landscape of agent memory research.
Toward Efficient Agents~\cite{yang2026toward} shifts the focus from architectural expressiveness to deployment cost, surveying efficiency-oriented techniques across three core agent components: memory, tool learning, and planning. In addition, this survey discusses compression, context management, and reinforcement-learning-based reward design as shared principles to reduce latency, token consumption, and interaction steps.
More recently, Rethinking Memory Mechanisms~\cite{huang2026rethinking} assembles a large-scale
survey of over 200 papers, organizing memory along three dimensions: substrate, cognitive mechanism, and subject, while
reviewing learning policies over memory operations and cataloguing existing evaluation benchmarks.
From Storage to Experience~\cite{luo2026storage} offers an evolutionary perspective, formalizing memory development into three progressive stages: storage, reflection, and
experience. It also identifies long-range
consistency, dynamic environments, and continual learning as the core drivers of this evolution.
Graph-based Agent Memory~\cite{yang2026graph} narrows the scope to graph-based memory paradigms of knowledge graphs, temporal graphs, hypergraphs, and hierarchical
trees. In addition, it systematically analyzes extraction, storage, retrieval, and evolution along the memory lifecycle.

\section{Prompt Library}
\label{app:prompts}
This section details the prompt templates used for all experimental evaluations. To ensure reproducibility, we provide the specific instructions for memory construction, query processing, and the varying sensitivities of our evaluation protocols.

To provide a structured comparison, we classify the prompt designs into three operational stages: Memory Construction (Build), Query Processing (Query), and Response Generation (Answer). Table~\ref{tab:prompt_taxonomy} summarizes the design patterns across the evaluated systems.

\begin{table*}[t]
\centering
\small
\caption{Taxonomy of System Operation Prompts across Memory Architectures.}
\label{tab:prompt_taxonomy}
\resizebox{\textwidth}{!}{
\begin{tabular}{l | l | l | l}
\toprule
\textbf{System} & \textbf{Build Strategy (Memory Construction)} & \textbf{Query Strategy} & \textbf{Answer Strategy (Synthesis)} \\
\midrule
\textbf{MemoryOS} & Profile-based (User profiling, Knowledge extraction) & N/A (Direct Semantic Search) & Role-playing \& Profile-enriched \\
\textbf{AMem}     & Flat/Turn-based (Content analysis)                   & LLM Keyword Extraction       & Retrieved Memory Context \\
\textbf{Nemori}   & Episodic (Boundary detection, Episode generation)    & N/A (Direct Semantic Search) & Episode-based Retrieval \\
\textbf{MAGMA}    & Graph-based (Event extraction, Multi-hop reasoning)  & Multi-hop Entity Parsing     & Graph Traversal Synthesis \\
\textbf{SimpleMem}& Minimalist/Turn-level                                & Keyword Generation           & Context-based \\
\bottomrule
\end{tabular}
}
\vspace{-10pt}
\end{table*}

\subsection{Memory Construction and Retrieval}
Different memory architectures require different construction strategies. This section outlines the prompts used by MAG systems to consolidate raw interaction history into long-term storage and refine user queries.

\paragraph{Build Prompts (Memory Indexing)}
Used by MAG systems to consolidate raw interaction history into long-term storage. The evaluated systems utilize distinct structural representations:
\begin{itemize}
    \item \textbf{Profile Based (MemoryOS):} Instructs the LLM to extract observable user traits and merge them into an evolving profile.
    \item \textbf{Episodic (Nemori):} Segments continuous dialogue into discrete episodes using boundary detection.
    \item \textbf{Graph Based (MAGMA):} Translates interactions into relational structures (e.g., event extraction, triplet arrays).
\end{itemize}

\begin{quote}
\itshape
[PLACEHOLDER: Insert Build Prompt for memory consolidation, e.g., "Summarize the following interaction into atomic facts..."]
\end{quote}

\paragraph{Query Refinement Prompts}
While many systems (like Nemori and MemoryOS) bypass explicit refinement in favor of direct semantic search algorithms, systems like AMem and SimpleMem use LLMs to transform user queries into optimized search vectors or keywords.
\begin{quote}
\itshape
[PLACEHOLDER: Insert Query Prompt, e.g., "Given the conversation history, rewrite the user query for better retrieval..."]
\end{quote}

\subsection{Response Generation}

\paragraph{Answer Generator Prompts}
The standard templates used by all baselines (RAG, MAG, and FullContext) to produce final responses based on retrieved or provided context. The prompt designs vary based on the context strategy (e.g., profile-enriched role-playing for MemoryOS, episodic retrieval for Nemori) and specific constraints (e.g., temporal awareness or multi-hop reasoning for AMem and MAGMA).
\begin{quote}
\itshape
[PLACEHOLDER: Insert Answer Generator Prompt, e.g., "You are an assistant with access to the following memory shards. Answer the question based on..."]
\end{quote}

\subsection{LLM-as-a-Judge Evaluation Protocols}
\label{sec:LLM-as-a-Judge}
We utilize \texttt{gpt-4o-mini} as our backbone judge. To comprehensively evaluate architecture performance across the diverse grading criteria mentioned in Section~\ref{subsubsec:promptsdiff}, we structure our evaluation into two categories: literature-derived baseline prompts and sensitivity rubrics.

\subsubsection{Literature-Derived Baselines}
These prompts represent different community standards for "correctness" and are sourced directly from existing benchmarks.

\paragraph{Prompt 1: MAGMA (Semantic Correctness \& Context)}
Derived from the MAGMA framework~\cite{jiang2026magma}, this multi-level scoring protocol prioritizes information integration and reasoning. It emphasizes interpersonal knowledge retrieval and semantic equivalence, with specific guidelines for temporal and factual preservation.
\begin{promptbox}
Score the answer on a scale from \textbf{0.0 to 1.0} based on semantic correctness.

\textbf{Scoring Scale:} \\
- \textbf{1.0}: Perfect match — contains all key information, semantically equivalent \\
- \textbf{0.8}: Mostly correct — captures main point but may have minor differences \\
- \textbf{0.6}: Partially correct — has some correct info but incomplete \\
- \textbf{0.4}: Somewhat related — touches on topic but misses significant info \\
- \textbf{0.2}: Barely related — answer is mostly incorrect \\
- \textbf{0.0}: Completely wrong — answer is unrelated or contradicts gold answer 

\textbf{Instruction:} \\
Focus on user-interpersonal knowledge and temporal generosity. \\
Focus on semantic equivalence, not exact wording. \\
Assign partial credit for partially correct answers. 

\textbf{Input:} \\
Question: \{question\} \\
Gold answer: \{gold\_answer\} \\
Generated answer: \{generated\_answer\} 

\textbf{Output (JSON):} \\
\{ "score": 1.0, "reasoning": "..." \}
\end{promptbox}

\paragraph{Prompt 2: Nemori (Generous Semantic Matching)}
Adapted from the Nemori paper~\cite{nan2025nemori}, this is a lenient, semantics-oriented evaluation scheme. It emphasizes entity recall and judges whether the generated answer captures the same underlying concept as the ground truth using a binary (\texttt{CORRECT}/\texttt{WRONG}) classification, tolerating paraphrasing and verbosity.

\begin{promptbox}
Your task is to label an answer as \textbf{CORRECT} or \textbf{WRONG}.

You will be given: \\
(1) a question \\
(2) a gold (ground truth) answer \\
(3) a generated answer 

\textbf{Evaluation Guidelines:} \\
- Be generous in grading. \\
- If the generated answer conveys the same meaning or topic as the gold answer, mark it as CORRECT. \\
- Ignore differences in wording, phrasing, or length. \\
- Accept paraphrases and semantically equivalent answers. \\
- For time-related questions, accept different formats (e.g., ``May 7'' vs ``7 May''). 

\textbf{Input:} \\
Question: \{question\} \\
Gold Answer: \{gold\_answer\} \\
Generated Answer: \{generated\_answer\} 

First, provide a one-sentence reasoning, then output the result. 

\textbf{Output (JSON):} \\
\{ "label": "CORRECT" \} \quad or \quad \{ "label": "WRONG" \}
\end{promptbox}

\paragraph{Prompt 3: SimpleMem (Relevance \& Accuracy)}
Adapted from the SimpleMem baseline~\cite{liu2026simplemem}, this prompt focuses on retrieval precision and core fact preservation. It explicitly balances relevance, factual grounding, and tolerance to representational variation via a \textit{Robustness Protocol}.

\begin{promptbox}
You are an expert \textbf{Relevance \& Accuracy Evaluator}.

Your task is to determine whether the Predicted Answer successfully retrieves the necessary information to answer the Question, based on the Reference Answer.

\textbf{Input:} \\
Question: \{question\} \\
Reference Answer: \{reference\} \\
Predicted Answer: \{prediction\} 

\textbf{Evaluation Criteria:} 

1. \textbf{Responsiveness to Query} \\
The predicted answer must directly address the specific question and remain topically aligned with the user's intent. 

2. \textbf{Core Fact Preservation} \\
The prediction must capture the \textit{key signal} or \textit{core entity} from the reference (e.g., who, what, or outcome). 

3. \textbf{Informational Utility} \\
The answer must provide meaningful value. Even if concise, it should convey the essential information required by the question. 

4. \textbf{Robustness Protocol (Acceptable Variances)} \\
You must treat the following variations as valid matches: \\
- Temporal \& numerical tolerance (e.g., $\pm$1--2 days, rounded numbers) \\
- Granularity differences (e.g., ``Afternoon'' vs.\ ``14:05'', ``Late October'' vs.\ ``Oct 25'') \\
- Information subsetting (partial but sufficient answers) \\
- Synonymy and format variation 

\textbf{Grading Logic:} \\
- \textbf{Score 1.0 (Pass):} Contains relevant core information OR satisfies robustness conditions above. \\
- \textbf{Score 0.0 (Fail):} Missing core information, irrelevant, or fails to answer the question. 

\textbf{Output Format (JSON only):} \\
\{ \\
\quad "score": 1.0, \\
\quad "reasoning": "Brief explanation focusing on relevance and core match." \\
\}
\end{promptbox}

\section{Baseline Configurations}
\label{app:baselines}

This section details the hyper-parameter settings and model versions for the evaluated architectures. To ensure fair and objective comparisons, we strictly follow the default configuration settings provided in their respective open-source repositories, with the following standardized modifications applied across all baseline systems:

\begin{itemize}
    \item \textbf{Embedding Model:} All dense retrieval operations are uniformly configured to use \texttt{all-MiniLM-L6-v2}~\cite{wang2020minilm}, replacing any system-specific default embedding models to ensure a controlled baseline for semantic matching.
    \item \textbf{LLM Temperature:} The generation temperature is fixed at $0.3$ across all backbone LLMs to maintain a consistent balance between determinism and reasoning capability.
    \item \textbf{Retrieval Top-$k$:} For final answer synthesis that relies on retrieving raw conversation history or memory chunks, we uniformly set the retrieval scope to top-$k=10$.
    \item \textbf{Max Tokens:} The maximum token limits for generation and context windows are maintained at their repository-specific defaults to respect the intended design of each architecture.
\end{itemize}

A summary of these unified hyper-parameters alongside the specific operational parameters for each evaluated system is provided in Table~\ref{tab:hyperparameters}.

\begin{table*}[t]
\centering
\small
\caption{Key hyper-parameter configurations for the evaluated memory architectures. To ensure fair comparison, embedding models, LLM generation temperatures, and final answer retrieval scopes are strictly standardized across all baselines, while structural capacity parameters (e.g., max tokens) follow repository defaults.}
\label{tab:hyperparameters}
\resizebox{\textwidth}{!}{
\begin{tabular}{l | c c c | l }
\toprule
\textbf{Method} & \textbf{Embedding Model} & \textbf{LLM Temp.} & \textbf{Final Answer Top-$k$} & \textbf{System-Specific Defaults (Max Tokens \& Structure)} \\
\midrule
\textbf{Full Context} & N/A & 0.3 & N/A & Max Tokens: 128k (\texttt{gpt-4o-mini}) \\
\textbf{LOCOMO} & \texttt{MiniLM-L6-v2} & 0.3 & 10 & Max Tokens: Default; Buffer Size: Default \\
\textbf{AMem} & \texttt{MiniLM-L6-v2} & 0.3 & 10 & Max Tokens: Default; Keyword Extractor Temp: Default \\
\textbf{MemoryOS} & \texttt{MiniLM-L6-v2} & 0.3 & 10 & Max Tokens: Default; Update Frequency: Default \\
\textbf{Nemori} & \texttt{MiniLM-L6-v2} & 0.3 & 10 & Max Tokens: Default; Boundary Temp: 0.1 \\
\textbf{MAGMA} & \texttt{MiniLM-L6-v2} & 0.3 & 10 & Max Tokens: Default; Consolidation Threshold: Default \\
\textbf{SimpleMem} & \texttt{MiniLM-L6-v2} & 0.3 & 10 & Max Tokens: Default; Synthesis Strategy: Default \\
\bottomrule
\end{tabular}
}
\vspace{-10pt}
\end{table*}

% ---------------------------------------------------------
% Table 1: methodology and models
% ---------------------------------------------------------
\begin{table*}[t]
\centering
\small
\caption{Overview of memory systems and experimental configurations. We use \texttt{gpt-4o-mini} as the primary controller for all methods in the main benchmark to normalize reasoning costs.}
\label{tab:baselines}
\setlength{\tabcolsep}{6pt}
\renewcommand{\arraystretch}{1.15}
\begin{tabular}{l p{4.2cm} p{4.6cm} p{4.0cm}}
\toprule
\textbf{Method} & \textbf{Memory Structure} & \textbf{Update Policy} & \textbf{Retrieval Scope} \\
\midrule

\textbf{A-MEM} 
& Linked Node Graph (Atomic Units + Tags) 
& Evolutionary: LLM-based node rewriting \& dynamic linking 
& Dense Embedding Similarity (Top-$k$) \\

\textbf{MemoryOS} 
& Hierarchical Tiers (STM $\rightarrow$ LTM) 
& Rule-based: Frequency/Recency-based promotion 
& Cascading Hierarchy Search \\

\textbf{MAGMA} 
& Multi-relational Graph + Vector Index 
& Asynchronous: Dual-stream consolidation (Long-term) 
& Intent-guided Subgraph Traversal \\

\textbf{Nemori} 
& Dual Memory (Episodic Tree + Semantic Graph) 
& Gradient-inspired: Contextual memory modification 
& Hybrid (Top-$k$ Episodes + Semantic Facts) \\

\textbf{SimpleMem} 
& Hybrid Index (Dense/Sparse) of Compressed Units 
& Synchronous: On-the-fly synthesis \& deduplication 
& Planner-guided Multi-view Search \\

\bottomrule
\end{tabular}
\end{table*}

\section{Case Studies: Why Lexical Metrics Fail}
\label{app:metric_cases}

To further investigate the ranking discrepancies observed in Section~\ref{subsubsec:promptsdiff}, we conduct a qualitative analysis of representative cases where lexical metrics (e.g., F1) disagree with semantic judgments. Rather than presenting isolated examples, we organize these cases into a set of recurring failure mechanisms that reflect inherent limitations of token level evaluation.

We identify four common patterns:

\begin{itemize}
    \item \textbf{Surface Variation:} Correct answers expressed with additional context or alternative phrasing are penalized due to reduced lexical overlap.
    \item \textbf{Semantic Equivalence Gap:} Equivalent meanings conveyed through different formats or synonyms result in zero or near-zero scores.
    \item \textbf{Polarity Flip:} Minor lexical changes (e.g., negation) invert the semantic meaning while preserving high token overlap.
    \item \textbf{Entity Drift:} Incorrect entities or values are substituted within otherwise similar sentence structures, leading to inflated lexical similarity despite factual errors.
\end{itemize}

Table~\ref{tab:metric_cases} presents representative examples illustrating these failure modes. These cases demonstrate that lexical metrics are not merely noisy, but systematically misaligned with the abstraction, normalization, and reasoning behaviors exhibited by modern LLM-based systems.

\begin{table*}[t]
\centering
\small
\caption{Mechanism Oriented Failure Cases of Lexical Metrics. Lexical scores (F1) are contrasted with semantic judgments to highlight systematic mismatches.}
\label{tab:metric_cases}
\resizebox{\textwidth}{!}{
\begin{tabular}{p{0.30\textwidth} | p{0.18\textwidth} | p{0.14\textwidth} | c c | p{0.26\textwidth}}
\toprule
\textbf{Query \& Gold Truth} & \textbf{Model Answer} & \textbf{Failure Type} & \textbf{F1} & \textbf{Judge} & \textbf{Analysis} \\
\midrule

\textbf{Q:} What is the duration of the event? \newline
\textbf{Gold:} 18 days
&
The total duration was 18 days.
& Surface Variation
& 0.50
& 1.00
& Additional phrasing lowers lexical precision despite identical meaning. \\

\midrule

\textbf{Q:} What time does the event start? \newline
\textbf{Gold:} 14:00
&
2 PM
& Semantic Equivalence Gap
& 0.00
& 1.00
& Equivalent time representations yield zero token overlap. \\

\midrule

\textbf{Q:} Describe the price level. \newline
\textbf{Gold:} cheap
&
inexpensive
& Semantic Equivalence Gap
& 0.00
& 1.00
& Synonym substitution is not captured by lexical matching. \\

\midrule

\textbf{Q:} Is the software compatible with Mac? \newline
\textbf{Gold:} compatible with Mac
&
\textbf{not} compatible with Mac
& Polarity Flip
& 0.857
& 0.00
& Negation reverses meaning while preserving token overlap. \\

\midrule

\textbf{Q:} Who completed the project? \newline
\textbf{Gold:} John completed the project
&
Sarah completed the project
& Entity Drift
& 0.75
& 0.00
& Incorrect entity maintains structure but changes semantics. \\

\midrule

\textbf{Q:} How many items were included? \newline
\textbf{Gold:} three items
&
five items
& Entity Drift
& 0.50
& 0.00
& Numerical substitution is partially rewarded due to shared tokens. \\

\bottomrule
\end{tabular}
}
\end{table*}

These observations complement the quantitative findings in Section~\ref{subsubsec:promptsdiff} and suggest a fundamental mismatch: lexical metrics operate on surface form alignment, whereas agentic systems increasingly rely on abstraction, normalization, and compositional reasoning. As a result, improvements in reasoning quality may not be reflected and can even be penalized by traditional token-based evaluation.

\end{document}